\documentclass[preprint]{article}
\usepackage{neurips_2025}
\usepackage[utf8]{inputenc}
\usepackage[T1]{fontenc}
\usepackage{hyperref}
\usepackage{url}
\usepackage{booktabs}
\usepackage{amsfonts}
\usepackage{nicefrac}
\usepackage{microtype}
\usepackage{xcolor}
\usepackage{amsmath}
\usepackage[normalem]{ulem}
\usepackage{graphicx}
\usepackage{enumitem}
\usepackage[T1]{fontenc}
\usepackage[utf8]{inputenc}
\usepackage{inconsolata}
\usepackage{amsmath}
\usepackage{amssymb}
\usepackage{balance}
\usepackage{blindtext}
\usepackage{colortbl}
\usepackage{float}
\usepackage{makecell}
\usepackage{multicol}
\usepackage{multirow}
\usepackage{marvosym}
\usepackage{listings}
\usepackage{tcolorbox}
\usepackage{stfloats}
\usepackage{subcaption}
\usepackage{bm}
\usepackage{longtable}
\usepackage{array}
\usepackage{colortbl}
\usepackage{siunitx}

\newcommand\footnoteONLYtext[1]{
    \let \mybackup \thefootnote
    \let \thefootnote \relax
    \footnotetext{#1}
    \let \thefootnote \mybackup
    \let \mybackup \imareallyundefinedcommand}

\title{Co-Saving: Resource Aware Multi-Agent Collaboration for Software Development}

\author{
\textbf{Rennai Qiu}{\footnotesize $^\bigstar$\textsuperscript{\textdagger}} \quad
\textbf{Chen Qian}{\footnotesize $^\clubsuit$\textsuperscript{\textdagger}} \quad
\textbf{Ran Li}{\footnotesize $^\bigstar$} \quad
\textbf{Yufan Dang}{\footnotesize $^\bigstar$} \quad
\textbf{Weize Chen}{\footnotesize $^\bigstar$} \quad \\
\textbf{Cheng Yang}{\footnotesize $^\spadesuit$} \quad
\textbf{Yingli Zhang}{\footnotesize $^\diamondsuit$} \quad
\textbf{Ye Tian}{\footnotesize $^\heartsuit$} \quad
\textbf{Xuantang Xiong}{\footnotesize $^\heartsuit$} \quad \\
\textbf{Lei Han}{\footnotesize $^\heartsuit$} \quad
\textbf{Zhiyuan Liu}{\footnotesize $^\bigstar$\textsuperscript{\Letter}} \quad
\textbf{Maosong Sun}{\footnotesize $^\bigstar$\textsuperscript{\Letter}} \quad \\
{\footnotesize $^\bigstar$}Tsinghua University \quad
{\footnotesize $^\clubsuit$}Shanghai Jiao Tong University \quad \\
{\footnotesize $^\spadesuit$}Beijing University of Posts and Telecommunications \quad \\
{\footnotesize $^\diamondsuit$}Siemens \quad
{\footnotesize $^\heartsuit$}Tencent Robotics X \quad \\
\href{qrn22@mails.tsinghua.edu.cn}{\texttt{qrn22@mails.tsinghua.edu.cn}} \quad
\href{qianc@sjtu.edu.cn}{\texttt{qianc@sjtu.edu.cn}} \\
\href{liuzy@tsinghua.edu.cn}{\texttt{liuzy@tsinghua.edu.cn}} \quad 
\href{sms@tsinghua.edu.cn}{\texttt{sms@tsinghua.edu.cn}}
}

\begin{document}

\maketitle

\footnoteONLYtext{$^\dagger$Equal Contribution.}
\footnoteONLYtext{$^{\text{\Letter}}$Corresponding Author.}

\begin{abstract}
Recent advancements in Large Language Models (LLMs) and autonomous agents have demonstrated remarkable capabilities across various domains. However, standalone agents frequently encounter limitations when handling complex tasks that demand extensive interactions and substantial computational resources. Although Multi-Agent Systems (MAS) alleviate some of these limitations through collaborative mechanisms like task decomposition, iterative communication, and role specialization, they typically remain resource-unaware, incurring significant inefficiencies due to high token consumption and excessive execution time. To address these limitations, we propose a resource-aware multi-agent system---Co-Saving (meaning that multiple agents collaboratively engage in resource-saving activities), which leverages experiential knowledge to enhance operational efficiency and solution quality. Our key innovation is the introduction of "shortcuts"—instructional transitions learned from historically successful trajectories—which allows to bypass redundant reasoning agents and expedite the collective problem-solving process. Experiments for software development tasks demonstrate significant advantages over existing methods. Specifically, compared to the state-of-the-art MAS ChatDev, our method achieves an average reduction of 50.85\% in token usage, and improves the overall code quality by 10.06\%.
\end{abstract}

\section{Introduction}
\label{sec:introduction}
In recent years, Large Language Models (LLMs) have achieved remarkable success in various domains, including text generation, code synthesis, and long context comprehension \cite{vaswani2017attention, NEURIPS2020_1457c0d6, bubeck2023sparks}. However, the inherent limitations of standalone LLMs become apparent when they confront complex tasks that extend beyond conversational interactions, often exhibiting behaviors that are not sufficiently robust or adaptive \cite{AutoGPT}. Research in autonomous agents recently has improved LLMs by empowering them with features such as contextual memory \cite{park2023generative}, multi-step planning \cite{wei2022chain} and utilization of external tools  \cite{schick2023toolformer}.

Although these enhanced agents represent a significant leap forward, the increasing complexity of many challenges often surpasses the capabilities of any agent. This necessitates a further evolution towards collaborative approaches, providing a strong motivation for the development of MAS. MAS collaborate through mechanisms such as role assignment, task decomposition, and iterative communication \cite{park2023generative, li2023camel, qian2023communicative}, forming a chat chain between agents and thus achieving sophisticated goals that would be intractable for a single agent. MAS evidently offers advantages in its superior modularity, allowing for specialized agent roles; enhanced scalability, enabling the distribution of tasks across numerous agents; and increased robustness, providing resilience through redundancy and collective problem solving. These benefits have led to notable advancements in complex scenarios such as collaborative software development \cite{mills1976software, GPTEngineer, qian2023communicative}, graphical user interface (GUI) automation \cite{DBLP:conf/aaai/ZhangMM0WT25}, social simulation \cite{park2023generative, wang2023humanoid, hua2023war}, game playing \cite{wang2023voyager, zhu2023ghost, wang2023avalons, gong2023mindagent} and scientific research \cite{huang2023benchmarking, LLM-Research-Feedback-2023}.

\begin{figure}[ht]
    \centering
    \includegraphics[width=1.0\linewidth]{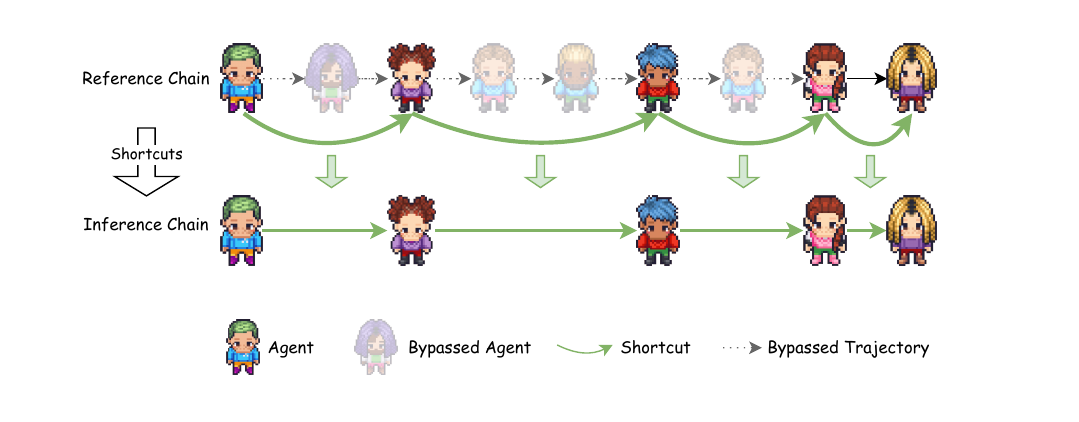}
    \caption{A schematic representation of the executing process, including reference chain and inference chain. The reference chain is based on historically excellent trajectories, while the inference chain is the execution process of the current task.}
    \label{fig:main-figure}
\end{figure}

However, MAS often confront unawareness \cite{zhang-etal-2024-timearena} of resources such as substantial token consumption and excessive time usage, which directly incurs inefficiency of system. As the scale of tasks expands and the number of participating agents increases, the frequency and complexity of agent interactions correspondingly increase, exacerbating operational overhead. Thus, effectively managing and reducing the operational overhead, while simultaneously enhancing resource efficiency, becomes imperative for the MAS. To address these limitations, we propose Co-Saving, a resource-aware multi-agent collaboration that leverages experiential knowledge to enhance both operational efficiency and solution quality. Our key innovation lies in introducing the concept of "shortcuts"—instructional transitions mined from historically successful trajectories. The shortcut serves as learned "fast track", enabling to effectively bypass redundant reasoning agents and accelerate problem-solving processes, particularly in familiar task contexts.

As the interaction process of agents in the MAS proceeds, a chat chain is accordingly formed, where the nodes correspond to the solutions generated by agents and the edges represent the instructions in the interaction process of agents. To fully utilize shortcuts to advance the current task execution process, a comprehensive evaluation for shortcuts is designed, involving both effectiveness and efficiency, and shortcuts filtering is implemented accordingly, which is shown schematically in Figure~\ref{fig:main-figure}. A force termination mechanism is also integrated to prevent resource exhaustion.

Experiments conducted on the SRDD dataset \cite{qian2023communicative} for software development tasks. Compared to baselines that including single-agent framework (e.g., GPT-Engineer \cite{GPTEngineer}) and existing multi-agent systems (e.g., MetaGPT \cite{hong2023metagpt}, ChatDev \cite{qian2023communicative}), our method achieves higher quality evaluated by co-learning \cite{qian-etal-2024-experiential} with lower cost. Specifically, compared to ChatDev, Co-Saving achieves an average reduction of 50.85\% in token consumption, along with a 10.06\% improvement of code overall quality.

\section{Method}
\label{sec:method}
In task-solving scenarios, particularly when addressing newly assigned tasks, it is often challenging to accurately estimate their inherent complexity or the resources required for successful completion by a multi-agent system (e.g., time, token consumption). To enhance the monitoring and management of task progress, we propose a strategy that involves retrieving reference tasks from historical records. These reference tasks function as a form of memory, guiding the agent in its current task execution. To leverage these references effectively, experiential knowledge is extracted from a repository of past tasks and integrated into the task-solving process. However, not all prior experiences are directly transferable or beneficial to the task at hand. Consequently, a critical step in this strategy is the evaluation and selection of relevant experiences, aimed at optimizing task execution efficiency.

\begin{figure}[h]
    \centering
    \includegraphics[width=1.0\linewidth]{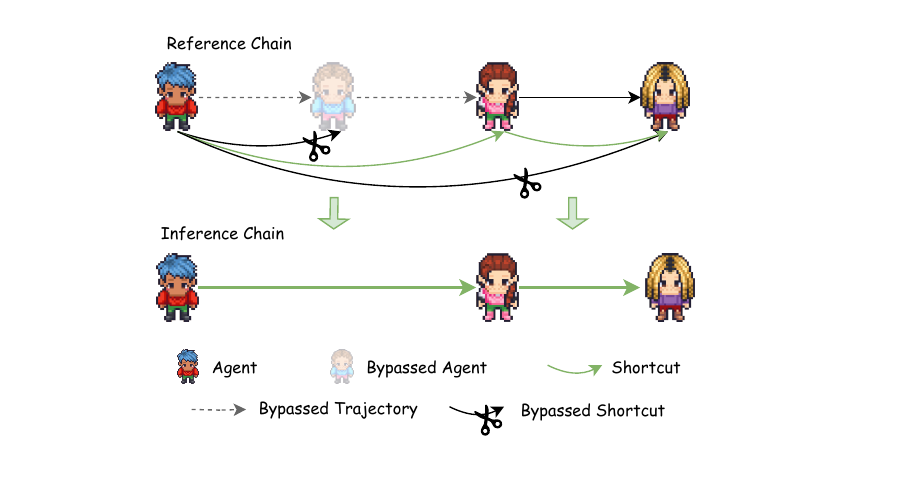}
    \caption{Overview of reference chain and inference chain to represent the shortcut filtering process. Being selected, evaluated and applied, shortcuts guide the current task to be completed in multiple steps.}
    \label{fig:reference-shortcut-graph}
\end{figure}

\subsection{Shortcut Formalization}

We introduce a type of instruction, termed a "shortcut", which connects two nodes within a reasoning chain by bypassing certain intermediate reasoning steps. This design aims to reduce the overall length of the reasoning chain, thereby enhancing reasoning efficiency. Figure \ref{fig:reference-shortcut-graph} shows a illustration of shortcut filtering. To validate the effectiveness of this approach, it is essential to conduct a comprehensive and quantitative evaluation of the shortcut mechanism.

To enable a more rigorous representation and analysis of the multi-agent collaboration process, we abstract each complete task execution as a directed graph. During the interaction, an instructor issues a series of instructions ($I=\{i_1,i_2,\cdots,i_n\}$) , and an assistant generates corresponding solutions as responses. Each instruction includes comments or feedback on the preceding solution, while each solution represents a complete software code snippet. Accordingly, the entire collaboration process can be formally represented by a directed graph $G=(N,E)$ , as defined below.

\begin{equation}
\begin{aligned}
N &= \{n_j|j=0,1,\cdots,n\} \\
E &= \{(n_j,i_{j+1},n_{j+1})|n_j,n_{j+1}\in N, i_{j+1}\in I\}
\end{aligned}
\end{equation}

Here, $N$ denotes the set of nodes, each corresponding to a solution state, with $n_0$ representing the initial state (typically an empty solution). $E$ represents the set of the edges corresponding to the instructions. Each edge connects two nodes and represents the transition from one solution $s_j$ to the modified one $s_{j+1}$, guided by the instruction $i_{j+1}$.

Here, $N$ denotes the set of nodes, each corresponding to a solution state, with $n_0$ representing the initial state (typically an empty solution). $E$ denotes the set of directed edges, where each edge represents an instruction guiding the transition from one solution $s_j$ to its subsequent modification $s_{j+1}$, based on the instruction $i_{j+1}$.

To enhance task completion efficiency, we aim for agents to achieve equivalent outcomes with fewer reasoning steps. For example, a solution that originally evolves through two steps (from $s_0$ to $s_1$ to $s_2$) could be optimized into a single-step transition (from $s_0$ directly to $s_2$). To this end, we introduce the concept of a shortcut, which is also modeled as a directed edge in graph $G$. A shortcut connects two non-adjacent nodes, always pointing forward in the interaction sequence, effectively bypassing intermediate reasoning steps while preserving the correctness of the final solution.

Let $S$ denote the set of all shortcuts, formally defined as follows:

\begin{equation}
S = \{(n_i, n_j)|n_i,n_j\in N, i<j\}
\end{equation}

We extract shortcuts from all tasks in the training set and store them in the form of instructions, serving as experiential knowledge. Subsequently, these shortcuts are incorporated into the agent's memory, allowing the agent to leverage prior experiences to enhance task-solving performance.

\subsection{Shortcut Filtering}

Not all extracted shortcuts are effective and efficient in improving solution generation or reducing resource consumption for a given task. Therefore, evaluating and selecting appropriate shortcuts is essential. We heuristically score shortcuts across multiple dimensions and ultimately derive a comprehensive metric to assess their overall utility.

Throughout the task execution process, we continuously monitor the current resource consumption, including time and token usage. When considering a shortcut, agents are guided to refer to its content and provide feedback accordingly, facilitating the optimization of candidate solutions. Shortcuts whose estimated resource consumption exceeds the remaining available resources are discarded from consideration. Only the feasible subset of shortcuts is retained for further evaluation. This selection process can be formalized as follows:
\begin{equation}
    S \leftarrow \{s|s \in S, t_s<t_r, \tau_s < \tau_r)\}
\end{equation}

where $t_s$ and $\tau_s$ denote the time and tokens required to generate the shortcut, respectively, while $t_r$ and $\tau_r$ represent the currently remaining time and tokens.

\paragraph{Value}

The contribution of a shortcut is primarily reflected in the transition it facilitates between two solutions—specifically, the transition from one node to another in the solution graph. For a given solution denoted by $n_j$ located at a specific node, we define its score as follows:
\begin{equation}
w(n_j) = \text{sim}(n_j, \text{task}) \times \text{sim}(n_j, s_{|N|}) \times [\![s_j]\!]
\end{equation}
Here, $s_{|N|}$ denotes the solution at the final node in the graph, representing the ultimate goal. The variable \textit{task} refers to the original software development requirement expressed in natural language. The two similarity terms are computed as the cosine similarity between the embedding vectors of the corresponding texts or code. The indicator function $[\![ \cdot ]\!]$ is binary: it equals 1 if the code corresponding to $s_j$ can be successfully compiled using an external compiler, and 0 otherwise.

Based on this node-level score, the value of a shortcut $(n_i, n_j)$ is defined as:
\begin{equation}
v(n_i, n_j) = w(n_j) - w(n_i)
\end{equation}
This value quantifies the incremental benefit that the shortcut brings to the software development process by enabling a more effective and efficient transition between solutions.

\paragraph{Cost}

Considering task-solving in multi-agent systems, the primary cost components are of two distinct types: time and tokens. These represent different dimensions of resource consumption and exhibit distinct distribution patterns within the dataset. To enable a unified evaluation, we normalize their raw values into percentile ranks based on their empirical distributions in the dataset. By integrating these normalized values, we derive a composite metric referred to as cost.

For a given shortcut $s_0$, let its normalized time and token consumption be $t_0$ and $\tau_0$, respectively. Denote by $T$ the set of normalized time values for all shortcuts in the dataset, denoted as $\mathcal{S}$, and by $\mathcal{T}$ the corresponding set of normalized token values.

We define the relative rankings of $s_0$ in terms of time and tokens as follows:
\begin{equation}
\alpha = \dfrac{|\{t<t_0|t\in T\}|}{|\mathcal{S}|}, \beta = \dfrac{|\{\tau<\tau_0|\tau \in \mathcal{T} \}|}{|\mathcal{S}|}
\end{equation}
The composite cost is then computed using the harmonic mean of $\alpha$ and $\beta$:
\begin{equation}
C = F_{\gamma}(\alpha, \beta) = \dfrac{2\alpha\beta}{\alpha+\beta}
\end{equation}
This formulation balances the trade-off between time and token efficiency, where $\gamma$ is the emergency factor that will be introduced in the following section.

\paragraph{Emergency Factor}

The value and cost metrics represent two distinct dimensions in evaluating task execution: value reflects the improvement in solution quality, while cost measures the efficiency of task completion. At different stages of task execution, the relative importance of these two aspects may vary. For instance, during the early stages—when resources are still abundant—the primary focus is typically on achieving high-quality solutions. Conversely, as resources approach depletion, the emphasis shifts toward completing the task promptly and within budget.

To accommodate these dynamic shifts in priority, we introduce the emergency factor $\gamma$, which regulates the relative weighting of value and cost throughout the task execution process. Unlike value and cost—which are determined solely by the characteristics of the shortcuts and the dataset—the emergency factor is explicitly linked to the user-defined resource budget, rendering it inherently dynamic and adaptive.

Let $t$ and $\tau$ denote the allocated budgets for time and tokens, respectively, and let $t_{\text{u}}$ and $\tau_{\text{u}}$ represent the corresponding amounts consumed thus far. The emergency factor $\gamma$ is then defined as follows:
\begin{equation}
\begin{aligned}
& \gamma_{t} := \frac{t_{\text{u}}}{t}, \gamma_{\tau} := \frac{\tau_{\text{u}}}{\tau}. \\
& \gamma = F_1(\gamma_{t}, \gamma_{\tau}) = \frac{2\gamma_{t}\gamma_{\tau}}{\gamma_{t} + \gamma_{\tau}}.
\end{aligned}
\end{equation}

\section{Experiments}
\label{sec:experiment}

\paragraph{Baselines}

To evaluate the effectiveness of our method, we select a diverse set of representative LLM-driven software engineering methods and pure LLMs to facilitate a comprehensive multidimensional comparison:

\begin{enumerate}[label=$\bullet$]
    \item GPT-3.5-Turbo \cite{ouyang2022training}, GPT-4 \cite{openai2024gpt4technicalreport}, LLaMA 3 70B \cite{touvron2023llama}and are widely adopted foundation models that serve as baselines for pure LLM performance, covering a range of capabilities from efficient instruction-following to strong multimodal reasoning and open-source adaptability.
    \item GPT-Engineer \cite{GPTEngineer} exemplifies a single-agent approach and serves as a foundational framework in this domain. Its key strength lies in its ability to interpret natural language requirements and autonomously perform development tasks such as code generation and execution through single step reasoning.
    \item ReAct \cite{yao2023reactsynergizingreasoningacting} integrates reasoning and acting within LLMs by jointly generating reasoning traces and environment-interacting actions. Unlike approaches that separate thought and execution, ReAct enables LLMs to iteratively refine their understanding and update the environment through interleaved reasoning and action steps.
    \item MetaGPT \cite{hong2023metagpt} adopts a MAS design, introducing a novel role-assignment mechanism where agents are assigned specific responsibilities. These agents collaborate through a standardized communication protocol to accomplish software engineering tasks.
    \item ChatDev \cite{qian2023communicative} presents a comprehensive multi-agent collaboration framework that decomposes the software development lifecycle into distinct phases, including demand analysis, code implementation, code review, and system testing. Within this framework, agents engage in multi-turn dialogues to iteratively propose instructions and solutions, thereby enhancing the quality and robustness of the software development process.
\end{enumerate}

\paragraph{Datasets}

We use a subset of the SRDD \cite{qian2023communicative} as our experimental corpus, containing diverse software development requirements. The dataset is organized into five primary categories: Education, Work, Life, Game, and Creation, and further divided into 40 fine-grained subcategories. We partition it into a training set for shortcut extraction and a test set for evaluation and data collection.

\paragraph{Metrics}

Our primary research objective is to enhance both the quality and efficiency of task completion in MAS, using software development as the application context. Accordingly, we evaluate task outcomes—specifically code generation—along two key dimensions: \textbf{quality} and \textbf{efficiency}.

For quality assessment, we adopt a comprehensive evaluation framework inspired by co-learning \cite{qian-etal-2024-experiential}, which integrates multiple dimensions into a unified metric for holistic evaluation. Efficiency is measured by the Budgeted Completion Rate (BCR), defined as the proportion of tasks completed within the specified resource constraints.
\begin{enumerate}[label=$\bullet$]

    \item \textbf{Completeness}: Measures whether the generated code provides a structurally complete implementation of the software requirement. It is quantified as the proportion of source files that do not contain placeholders such as "\texttt{TODO}".
    
    \item \textbf{Executability}: Assesses the ability of the generated software to compile and run successfully in a real operating system environment. It is calculated as the ratio of programs that compile and execute without errors.
    
    \item \textbf{Consistency}: Evaluates the semantic alignment between the generated code and the original natural language requirement, computed as the cosine similarity between their respective embedding vectors.
    
    \item \textbf{Granularity}: Assesses the level of detail in the generated code. Given the inherent challenges in objectively quantifying code granularity and completeness—especially across tasks of varying complexity—we adopt the average number of lines of code per task as a practical proxy. A higher value indicates greater code detail.
    
    \item \textbf{Quality}: A comprehensive metric obtained by integrating completeness, executability, consistency, and granularity. Specifically, it is defined as the product of these four metrics, serving as an overall indicator of code quality.
    
    \item \textbf{Budgeted Completion Rate} (\textbf{BCR}): Measures the proportion of tasks completed within the predefined resource budget (time and tokens). It reflects resource efficiency without considering the quality of the generated solution; thus, even low-quality code produced quickly is counted under this metric.

\end{enumerate}

\paragraph{Implementation Details}

The software development process is divided into multiple phases, including demand analysis, language selection, code completion, code review, and system testing. Our work primarily focuses on phases directly related to code generation. For these tasks, we adopt GPT-3.5-Turbo as the base model. For node evaluation, metric consistency computation, and reference task retrieval, we employ text-embedding-ada-002 as the semantic embedder, due to its strong performance in both textual and code-related embeddings. Python 3.9.19 serves as the external feedback environment, enabling compilation, execution, and assessment of generated code.Throughout the experiments, we monitor agent interactions and implicitly construct the interaction graph. The number of edges in the graph corresponds to the number of interaction rounds. To prevent excessive interactions, once the current interaction graph reaches or exceeds the number of edges in the reference task graph, we forcibly terminate the task.

\begin{table}[ht]
\centering
\caption{overall performance of selected baseline and our Co-Saving. The \textbf{highest} scores are formatted in bold and the \uline{second-highest} scores are underlined.}
\label{tab:main-results}
\resizebox{0.95\textwidth}{!}{
\begin{tabular}{lcccccc}
\toprule
Method        & Completeness & Executability & Consistency & Granularity & Quality & BCR \\
\midrule
GPT-3.5-Turbo &0.9200         &0.8600         &\textbf{0.8076}&0.2882         &0.1842         &\textbf{1.0000} \\
GPT-4          &\textbf{0.9800}&\uline{0.8800} &\uline{0.8053} &0.3036         &\uline{0.2109} &0.6200 \\
LLaMA 3 70B    &0.9000         &0.7600         &0.8032         &0.4333         &0.2058         &\textbf{1.0000} \\
\midrule
GPT-Engineer &0.5200          &0.5680         &0.7863         &0.3379         &0.0785         &\textbf{1.0000} \\
ReAct        &\uline{0.9600}    &\textbf{1.0000}&0.8033         &0.2548         &0.1965         &\textbf{1.0000} \\
\midrule
MetaGPT      &0.7040          &0.1120         &0.7731         &\textbf{0.9721}&0.0593         &0.2720 \\
ChatDev      &0.9040          &0.3680         &0.7897         &\uline{0.5746} &0.1510         &0.0160 \\
\midrule
Co-Saving           &0.8160          &0.6880         &0.8034         &0.5743         &\textbf{0.2515}&\uline{0.7280} \\
\bottomrule
\end{tabular}
}
\end{table}

\subsection{Overall Performance}

As shown in Table \ref{tab:main-results}, our proposed approach (denoted as Co-Saving\footnote{Co-Saving means that multiple agents collaboratively engage in resource-saving activities.}) significantly outperforms all baselines in terms of Quality and surpasses other multi-agent baselines in BCR. These results indicate that Co-Saving effectively accelerates the reasoning trajectory toward generating high-quality solutions.

As single-agent frameworks, GPT-Engineer and ReAct typically do not decompose or subdivide tasks based on user instructions. Instead, they perform code generation through a one-shot reasoning process. Consequently, they exhibit low execution time and resource consumption. The same observation holds for pure LLM-based paradigms. However, for more complex software development tasks, these approaches often fail to produce functionally complete code. In many cases, they define interfaces or modules related to complex requirements but leave them partially or entirely unimplemented. This limitation artificially inflates the Executability metric, as syntactically correct but semantically incomplete code can still compile and run. Such shortcomings are reflected in the relatively low Granularity scores, which indicate insufficient implementation detail.

In contrast, ChatDev adopts a multi-stage reasoning paradigm that iteratively refines solutions, leading to more complete implementations. However, this iterative process incurs higher resource consumption, resulting in a lower BCR. MetaGPT achieves a BCR between GPT-Engineer and ChatDev. It leverages multi-agent collaboration through role-based coordination to perform multi-step reasoning, but still struggles to generate logically coherent code for complex tasks, leading to a relatively lower Executability score.

For the Completeness metric, ChatDev slightly outperforms Co-Saving. We hypothesize that this advantage stems from Co-Saving’s resource-awareness and dynamic execution control. When encountering tasks that exceed the available resource budget, Co-Saving may opt to terminate reasoning prematurely, prioritizing efficiency over completeness. In contrast, ChatDev lacks such resource sensitivity and continues execution regardless of task complexity, achieving higher completeness at the expense of increased resource usage.

Additionally, Consistency scores across all four experimental settings show only minor differences, with Co-Saving achieving a modest improvement. This result may reflect the limitations of current embedding models in capturing fine-grained semantic distinctions between code and textual requirements. Consequently, these models are insufficiently sensitive to subtle inconsistencies, highlighting the need for more precise evaluation methods to better assess code-text alignment.

\subsection{Ablation Study}
\label{sec:ablation}

In the Method section, we introduced key components of our approach: shortcut selection, cost design, and the emergency factor. To validate the effectiveness of each component, we design corresponding ablation studies. The results of the full model and the ablation variants are summarized in Table \ref{tab:ablation}.

\begin{table}[h]
\centering
\caption{Ablation study on main design in Co-Saving, $\backslash$ denote the removing operation, the three ablations remove selection, cost, emergency factor($\gamma$) respectively.}
\label{tab:ablation}
\resizebox{0.95\textwidth}{!}{
\begin{tabular}{lcccccc}
\toprule
Method                      & Completeness  & Executability & Consistency   &  Granularity  & Quality        & BCR \\
\midrule
Co-Saving                          &\textbf{0.9250}&\uline{0.8500} &\textbf{0.8106}&0.8556         &\uline{0.5453} &\uline{0.8000} \\
\quad $\backslash$ selection   &\uline{0.8500} &0.8000         &0.8073         &\uline{0.8791} &0.4826         &0.6000 \\
\quad $\backslash$ cost     &0.8250         &\textbf{0.9250}&0.8061         &\textbf{0.9412}&\textbf{0.5789}&0.7500 \\
\quad $\backslash$ $\gamma$ &0.8250         &\uline{0.8500} &\uline{0.8087} &0.8484         &0.4811         &\textbf{0.8250} \\
\bottomrule
\end{tabular}
}
\end{table}

As we can see, removing the cost-based shortcut selection mechanism results in all candidate shortcuts being retained for evaluation, including those that significantly exceed the available resource budget. Consequently, this variant exhibits a substantially lower BCR compared to other configurations. In the second ablation, where cost is removed from the value-cost evaluation metric (i.e., only value is considered), the system achieves relatively good performance in Executability and Granularity. However, the lack of resource awareness makes it difficult to complete tasks within time constraints, leading to lower Completeness and a reduced BCR. In the third ablation, the emergency factor is excluded. Without this dynamic adjustment, the system continues to prioritize high-value shortcuts even under resource-limited conditions. Although the BCR remains relatively high due to the forced termination mechanism, both Completeness and Granularity are lower compared to the full Co-Saving configuration, indicating suboptimal task outcomes.
 
\subsection{Resource Distribution Shift}

\begin{figure}
    \centering
    \includegraphics[width=1.0\linewidth]{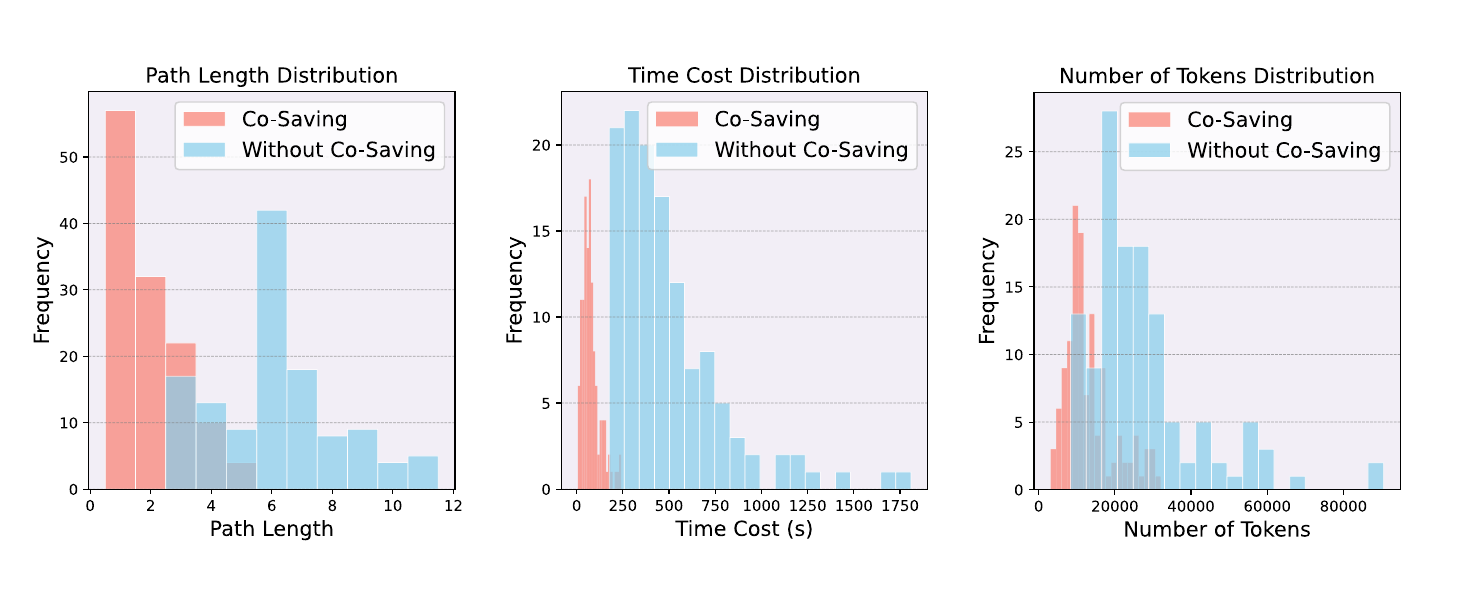}
    \caption{Distribution of path length, time cost and number of tokens. experiments with Co-Saving and without Co-Saving are indicated in red and blue color, respectively, as shown in legend.}
    \label{fig:distribution-figure}
\end{figure}

To further evaluate the effectiveness of Co-Saving, we conducted a comparative study between software development MAS with and without Co-Saving. Specifically, we analyzed the distribution of path lengths—defined as the number of edges in the execution graph, reflecting the number of reasoning iterations—on the same dataset. Additionally, we examined the distribution of resource consumption, including execution time and token usage. The experimental results are presented in Figure \ref{fig:distribution-figure}.

The inclusion of the Co-Saving algorithm results in a significant reduction in the number of reasoning iterations required for task execution. Additionally, both total execution time and token consumption are notably decreased. These findings demonstrate that Co-Saving effectively streamlines the multi-agent reasoning process, accelerating task execution and enhancing overall development efficiency.

This improvement is largely attributed to Co-Saving’s ability to accurately assess and utilize shortcuts. By extracting precise and efficient instructions from reference tasks, Co-Saving enables agents to make more informed decisions, thereby reducing the occurrence of inefficient or ineffective actions.

\subsection{Case Study}
\label{sec:case_study}

In order to illustrate how Co-Saving operates within a MAS, we present a case study of a specific task. Using ChatDev as the underlying software development framework, we select the task \emph{"Photo Defogger"} as an example. At the initial stage, the system retrieves the reference task \emph{"Background Blur Editor"} from the training dataset. This reference task forms an execution graph comprising three rounds of reasoning.

For the current task, after the programmer generates node $n_0$ in the \emph{Code Complete} stage, our system evaluates the shortcuts $(n_0, n_1)$, $(n_0, n_2)$, and $(n_0, n_3)$ to select the optimal path. Eventually, $(n_0, n_2)$ is chosen.In the reference task, the transition from $n_0$ to $n_1$ involves fixing a function to prevent file overwrite issues and adding necessary import statements. For the current task, given the programmer’s initial code and the shortcut $(n_0, n_1)$ as input, the code reviewer generates an instruction to adjust function details to avoid file overwrites. Based on this instruction, the programmer produces a new solution, corresponding to node $n_2$ in the reference task. It is worth noting that the shortcut $(n_0,n_2)$ is not a simple merge of the edges $(n_0,i_1,n_1)$ and $(n_1,i_2,n_2)$, but is related to $n_0$ and $n_2$, containing more complete and detailed information about how to transition from the source to the target. For instance, a shortcut says "To transition from the initial code version to the final version, follow these instructions: Modules and Classes: 1. In the game.py file, add the following import statement at the top... Data Structure: 1. In the player.py file, add the following attribute to the Player class... Main Program Flow: 1. In the game.py file, modify the  take\_turn method as follows... Exception Handling...". Without the shortcut input, the code reviewer could simply output short, abbreviated feedback.

Next, the shortcut originating from $n_2$ in the reference task—specifically, $(n_2, n_3)$—is considered. After evaluation, this shortcut is selected for code review, leading to the generation of another solution in the subsequent code modification stage. At this point, the number of reasoning steps in the current task reaches the predefined limit (matching the reference task’s path length), prompting termination of further inference.The execution processes of both the current and reference tasks are illustrated in Figure \ref{fig:reference-shortcut-graph}. Ultimately, Co-Saving successfully generates an executable program with a correct GUI interface and essential functions within three iterations. In contrast, ChatDev requires more iterations to produce a comparable solution, incurring higher token consumption.

\section{Related Work}
\label{sec:related-work}

Understanding and processing natural language remains a central challenge in artificial intelligence. LLMs \cite{NEURIPS2020_1457c0d6, vaswani2017attention, bubeck2023sparks, radford2019language, touvron2023llama, wei2022emergent, Shanahan2023, chen2021evaluating, brants2007large, ouyang2022training, yang2023large, qin2023large, kaplan2020scaling}, empowered by large-scale pretraining and parameter-rich architectures, have achieved remarkable advancements in this area. With the rapid development of LLMs, there is increasing interest in building autonomous agents \cite{zhou2023webarena, wang2023voyager, park2023generative, wang2023humanoid, wang2023promptagent, AutoGPT, GPTEngineer} that leverage LLMs for domain-specific tasks. These agents combine LLMs’ reasoning and language understanding capabilities with external tools \cite{schick2023toolformer, cai2023large, qin2023toolllm, ruan2023tptu, GPT4Tools}, context memory management \cite{park2023generative, sumers2024cognitive}, and task decomposition and planning strategies \cite{GPTEngineer, chen2023agentverse, liu2023bolaa, wei2022chain}, enabling them to tackle increasingly complex problems \cite{zhao2023expel, zhou2023webarena, ma2023laser, zhang2023generative, wang2023large, ding2023designgpt, weng2023prompt}. In parallel, techniques such as self-evolving \cite{hu2025selfevolving}, self-instruct \cite{wang-etal-2023-self-instruct}, and other enhancement methods \cite{fu2025agentrefine, lingam2025enhancing, chen2025internet, DBLP:conf/aaai/YueWCHW25, qian2025scaling, shi2025monte, DBLP:conf/aaai/Tabarsi25} have been proposed to further improve agent capabilities.Beyond single-agent research, MAS have emerged as a critical area of study \cite{hong2023metagpt, hua2023war, wu2023autogen, liang2023encouraging, chen2023agentverse, chan2023chateval, chen2023gamegpt, ding2023designgpt, guo2024large}. Unlike single-agent frameworks, which attempt to solve complex problems independently, MAS introduce greater variability and design flexibility. This includes assigning distinct roles and identities to different agents, designing workflows for decomposing complex tasks into subtasks, and establishing communication protocols, information exchange pathways, and coordination structures to facilitate collaborative task execution.

Recent studies have explored how the number and structure of agents influence the performance and scalability of MAS \cite{qian2025scaling}. As agent count and task complexity increase, interaction frequency and resource consumption also grow. This highlights key challenges in enhancing resource utilization, minimizing redundant communication, and designing efficient collaboration mechanisms. For instance, AgentDropout \cite{wang2025agentdropoutdynamicagentelimination} improves communication efficiency by pruning redundant agents and interactions in multi-round dialogues, enhancing token efficiency and task performance. BTP (Budget-Constrained Tool Learning with Planning) \cite{zheng-etal-2024-budget} formulates budget-aware tool selection strategies to maximize utility under resource constraints. TimeArena \cite{zhang-etal-2024-timearena} provides a simulated environment with complex temporal dynamics, revealing that current LLMs lack robust temporal reasoning, especially in multitasking or concurrent scenarios—underscoring the need for more temporally-aware agent designs.

\section{Conclusion}

\label{sec:conclusion}
In this paper, we proposed Co-Saving, a resource-aware multi-agent collaboration designed to address the inherent lack of resource sensitivity in multi-agent collaboration. By introducing shortcuts mined from successful historical trajectories, our system enables agents to leverage prior experience to optimize resource usage and accelerate task completion. The framework adopts a graph-based representation of task execution, where shortcuts connect effective reasoning paths to support efficient navigation through familiar problem-solving contexts. Additionally, we incorporate an emergency factor for dynamic shortcut management and a forced termination mechanism to prevent resource exhaustion. Experimental results on the SRDD demonstrate that Co-Saving significantly improves both efficiency and solution quality compared to baseline single-agent and multi-agent. Overall, this work underscores the importance of resource-aware multi-agent collaboration, offering a scalable and effective paradigm for deploying LLM-powered MAS in complex real-world applications.

\bibliographystyle{unsrt}
\bibliography{references}
\medskip
\end{document}